\documentclass[acmlarge]{acmart}
\AtBeginDocument{%
  \providecommand\BibTeX{{%
    \normalfont B\kern-0.5em{\scshape i\kern-0.25em b}\kern-0.8em\TeX}}}

\setcopyright{acmcopyright}
\copyrightyear{2018}
\acmYear{2018}
\acmDOI{XXXXXXX.XXXXXXX}

\acmConference[India HCI '23]{}{November 23--25,
  2023}{Dehradun, India}
\acmBooktitle{India HCI 2023,
 November 23--25, 2023, Dehradun, India} 
\acmPrice{15.00}
\acmISBN{978-1-4503-XXXX-X/18/06}

\usepackage{xcolor}
\usepackage{soul, color}
\usepackage{wrapfig}
\begin{document}

\title{
Exploring Social Media Posts for Depression Identification: A Study on Reddit.}

\author{Nandigramam Sai Harshit}
\email{nandigramam19@iiserb.ac.in}
\author{Nilesh Kumar Sahu}
\email{nilesh21@iiserb.ac.in}
\author{Haroon R. Lone}
\email{haroon@iiserb.ac.in}
\affiliation{%
  \institution{Indian Institute of Science Education and Research, Bhopal}
  \city{Bhopal}
  \state{Madhya Pradesh}
  \country{India}
  \postcode{462066}
}


\begin{abstract}


Depression is one of the most common mental disorders affecting an individual's personal and professional life.
In this work, we investigated the possibility of utilizing social media posts to identify depression in individuals. To achieve this goal, we conducted a preliminary study where we extracted and analyzed the top Reddit posts made in 2022 from depression-related forums. The collected data were labeled as depressive and non-depressive using UMLS Metathesaurus. Further, the pre-processed data were fed to classical machine learning models, where we achieved an accuracy of 92.28\% in predicting the depressive and non-depressive posts.

\end{abstract}

\ccsdesc[500]{Applied computing~Psychology}
\ccsdesc[500]{Applied computing~Decision analysis}
\ccsdesc[300]{Computing methodologies~Classification and regression trees}

\keywords{Mental health, Mental disorder, mHealth, Social networking, Depression, Classification, Reddit}


\maketitle

\section{Introduction}
A mental health disorder is characterized by clinically significant disturbance in an individual's cognition, emotion, or behavior. According to Global Health Data Exchange (GHDx),  one in every eight people suffered from mental disorders in 2019\footnote{\url{https://www.who.int/news-room/fact-sheets/detail/mental-disorders}}. Anxiety and depression are the most common mental health disorders and need utmost attention. 

Depression\footnote{\url{https://www.who.int/news-room/fact-sheets/detail/depression}} is characterized by persistent sadness and a lack of interest or pleasure in previously rewarding or enjoyable activities. Furthermore, the symptoms include changes in appetite, severe weight loss/gain, trouble sleeping, thoughts of death or suicide, and many other factors which may affect an individual's quality of life.


Various technologies like body sensors, electroencephalogram (EEG), smartphones, and wearables are being explored to identify early signs of depression \cite{cai2018pervasive, wang2018tracking}. Nowadays, social media posts are being actively explored for identifying and monitoring several disorders passively \cite{skaik2020using}. For example,  Wang et al.~\cite{wang2017detecting}  and Choudhary et al.~\cite{de2013predicting} used Twitter posts to predict eating disorders and depression, respectively. An article by Mude et al.~\cite{mude2023social} highlights that approximately 50\% of the Indian population engages with various social media platforms.
This increase in social media users provides a unique opportunity to investigate user-generated posts as a potential means of predicting depressive symptoms while simultaneously upholding user privacy.

This work showcases our preliminary findings of identifying people suffering from depression from Reddit posts. We focus on extracting user posts from two sub-Reddits, ``depression'' and ``happy''.
In both sub-Reddits, users share their daily life experiences, activities, and thoughts for discussion with other users. Using this data, we classified the posts/text as ``depressive'' or ``non-depressive'' by using UMLS (Unified Medical Language System) Metathesaurus\cite{UMLS}. 
UMLS Metathesaurus comprises a pre-compiled dictionary of words that classifies text based on keywords. 
We used several machine learning techniques for classification \& achieved an accuracy of 92.28\% with the Random Forest model.

\section{Dataset}

We searched various social media platforms (LinkedIn, Twitter, Reddit, Facebook) for creating a dataset. 
Among the different platforms, we found Reddit to be an apt data source. We used the PushShift API\footnote{\url{https://pypi.org/project/pushshift.py/}} for  extracting posts from several subreddits. The API allows to extract posts from a particular subreddit during a specified period.
We extracted data from subreddits closely related to depression, such as r/depression, r/ADHD, and r/psychology, for the year 2022. 
Further, we used the UMLS Metathesaurus, which analyses a paragraph or text and then compares the words in it with a pre-complied dictionary, and tells whether the paragraph has words which are related to ``depression'' corpus in the Metathesaurus. The Metathesaurus provides keywords that are associated with depression, such as ``sad'', ``tired'', and ``hopeless''. If these keywords are present in a sentence, it is more likely that the sentence is from someone who is depressed. For example, the following sentences contain the related keywords,
\begin{itemize}
    \item I feel lost. I'm \textbf{``unhappy''} it feels like that's never gonna change.
    \item   I feel \textbf{``worthless''} and \textbf{``hopeless''} and \textbf{``lonely''} and just \textbf{``miserable''}.
\end{itemize}

The data collected consists of text from the top (most upvoted/commented on) posts in subreddits.  We collected 1441 posts labeled as ``depression'' and 1165 posts as ``non-depression''. The posts collected from r/Happy were extracted to establish the ground truth data for ``non-depressed'' label.

\section{Methodology}

\subsection{Data Pre-processing}

\begin{figure}[htb]
  \centering
  \includegraphics[width=0.6\textwidth]{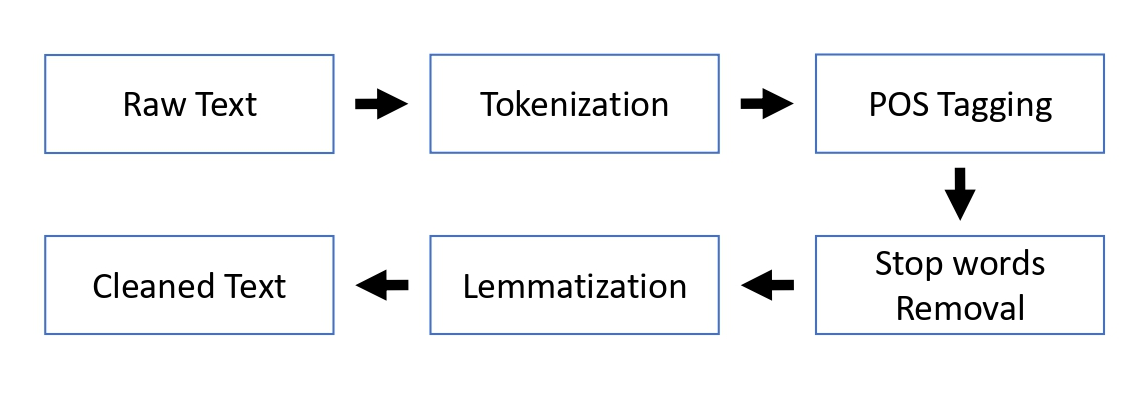} 
  \caption{Text pre-processing Pipeline} 
  \label{fig:pre-processing} 
\end{figure}

Data preprocessing involves tokenization, POS (Part-of-Speech) tagging, removing stopwords, and applying lemmatization, as shown in Figure~\ref{fig:pre-processing}. In tokenization, sentences are broken down into individual words known as tokens. Then all the tokens are converted to lowercase to ensure that words with the same spelling but different capitalization are treated as the same token, followed by grammatical (POS) tagging. Next, stop words such as `a', `an', `the', `in', and `of' are removed from the tokens, as these do not contribute to the meaning of a sentence followed by stemming/lemmatization. Stemming involves removing suffixes from words to get their stem (e.g., ``running'' becomes ``run''), while lemmatization involves using a dictionary to convert words to their base form (e.g., ``running'' becomes ``run'' or ``ran''). Stemming is a simpler method, while lemmatization produces more accurate results. 

\subsection{Classification}

We used the ``Bag Of Words'' (BOW) model to convert text data into statistical representations for machine learning algorithms. In BOW, initially a vocabulary is created by listing all unique words from the entire corpus (data collection). Each unique word becomes a ``token'' or a ``feature''. Then, for each document in the corpus, the frequency of each word in the vocabulary is computed. This creates a vector representation where each element corresponds to the word count in the vocabulary. Each document is represented as a vector, where the dimensions correspond to the words in the vocabulary, and the values represent the frequency of each word in the document.

We split the labeled dataset into training and testing, where 80\% of the data were used for training and 20\% of the data for testing the algorithms. Further, we explored Logistic Regression, Naive Bayes, Support Vector Machine (SVM), and Random Forest classifiers with default parameter settings. These models had varying degrees of accuracy.

\section{Results }

\begin{wraptable}{r}{0.44\textwidth}
\centering
\caption{Classification accuracy of different models}
\begin{tabular}{ lc  }
\toprule
\textbf{Model}& \textbf{Accuracy (\%)} \\
\midrule
Logistic Regression   & 90.39 \\
Naive Bayes&   86.27 \\
SVM& 92.10 \\
Random Forest& 92.28 \\
\bottomrule
\end{tabular}
\label{table:accuracy}
\end{wraptable}

Table \ref{table:accuracy} presents the accuracy of our classification models. Notably, the Random Forest algorithm yielded the highest accuracy at 92.28\%. This can be attributed to the proficiency of the algorithm in capturing complex data relationships and distinctive feature selection methods. However, we must note that the dataset currently used is small, and this could lead to the model predicting the wrong classes. We will overcome the drawback by increasing the dataset size, which will help provide more training data for the model.


\section{Conclusion and Future work}
In this preliminary work, we have demonstrated the potential use of social media posts to predict depression. Further, this work can be extended to larger datasets and social media sites such as Twitter and Facebook to get more user data. Social media analysis will also help us to understand how depression affects people.

\bibliographystyle{unsrt} 
\bibliography{sample-base}

\appendix
\end{document}